\documentclass[10pt,twocolumn,letterpaper]{article}

\usepackage{iccv}
\usepackage{times}
\usepackage{epsfig}
\usepackage{graphicx}
\usepackage{amsmath}
\usepackage{amssymb}
\usepackage{amsfonts}
\usepackage{sistyle}
\usepackage{subfigure}
\usepackage{comment}
\usepackage{algorithmic, algorithm}

\usepackage[pagebackref=true,breaklinks=true,letterpaper=true,colorlinks,bookmarks=false]{hyperref}

\iccvfinalcopy 

\def\iccvPaperID{2782} 

\ificcvfinal\fi

\newcommand{\argmin}{\mathop{\rm arg~min}\limits}
\newcommand{\argmax}{\mathop{\rm arg~max}\limits}
\newcommand{\bhline}[1]{\noalign{\hrule height #1}}

\begin{document}

\title{MSR-DARTS: Minimum Stable Rank of Differentiable Architecture Search}

\author{Kengo Machida\textsuperscript{\rm 1}, Kuniaki Uto\textsuperscript{\rm 1}, Koichi Shinoda\textsuperscript{\rm 1} and Taiji Suzuki\textsuperscript{\rm 2,3}\\
\textsuperscript{\rm 1}School of Computing, Tokyo Institute of Technology, Japan\\
\textsuperscript{\rm 2}Graduate School of Information Science and Technology, The University of Tokyo, Japan\\
\textsuperscript{\rm 3}Center for Advanced Intelligence Project, RIKEN, Japan\\
{\tt\small \{machida, uto\}@ks.c.titech.ac.jp, shinoda@titech.ac.jp, taiji@mist.i.u-tokyo.ac.jp}
}

\maketitle
\ificcvfinal\thispagestyle{empty}\fi

\begin{abstract}
   In neural architecture search (NAS), differentiable architecture search (DARTS) has recently attracted much attention due to its high efficiency.
   It defines an over-parameterized network with mixed edges, each of which represents all operator candidates, and jointly optimizes the weights of the network and its architecture in an alternating manner.
   However, this method finds a model with the weights converging faster than the others, and such a model with fastest convergence often leads to overfitting.
   Accordingly, the resulting model cannot always be well-generalized.
   To overcome this problem, we propose a method called minimum stable rank DARTS (MSR-DARTS), for finding a model with the {\it best generalization error} by replacing architecture optimization with the selection process using the minimum stable rank criterion.
   Specifically, a convolution operator is represented by a matrix, and MSR-DARTS selects the one with the smallest stable rank. 
   We evaluated MSR-DARTS on CIFAR-10 and ImageNet datasets. 
   It achieves an error rate of $2.54\%$ with $4.0$M parameters within $0.3$ GPU-days on CIFAR-10, and a top-1 error rate of $23.9\%$ on ImageNet. 
   The official code is available at \url{https://github.com/mtaecchhi/msrdarts.git}.
\end{abstract}

\section{Introduction}
Neural architecture search (NAS) seeks to design neural network structures automatically and has been successful in many tasks
\cite{ahn2018fast,liu2019auto,DBLP:conf/icml/PhamGZLD18}.
In NAS, all possible architectures are defined by a {\it search space}, which consists of network topologies and operator sets, and a {\it search strategy} is used to efficiently obtain a better architecture on the defined search space.
A recent trend in the search space is a small component in a network called a {\it cell}, which is defined as an optimization target to reduce search cost.
Reinforcement learning (RL) \cite{DBLP:conf/iclr/ZophL17,zoph2018learning,DBLP:conf/icml/PhamGZLD18} and evolutionary algorithms (EAs) \cite{DBLP:conf/iclr/LiuSVFK18,DBLP:conf/aaai/RealAHL19,DBLP:conf/icml/TangGD17} are widely used for the search strategy.

Recently, DARTS \cite{liu2018darts} and DARTS-based methods \cite{DBLP:conf/iclr/XieZLL19,chen2019progressive,xu2019pc,liang2019darts+} have been proposed, which are differentiable methods that relax the search spaces to be continuous, enabling direct application of gradient-based optimization.
These methods are effective regarding search cost since they skip the evaluation of each sampled architecture, which is required in RL and EAs.
The cell defined in the above studies is a direct acyclic graph (DAG) with multiple nodes, each of which is a latent representation ({\it e.g.}, a feature map in convolutional networks), and each directed edge is associated with an operator.
While these studies explicitly introduce {\it architecture parameters} as learnable parameters in addition to the weight parameters of  over-parameterized networks in the architecture search, each edge in a DAG is a {\it mixed edge} that includes all candidate operators in the operator set, and each operator is weighted by an architecture parameter.
An architecture parameter indicates how suitable its operator is in a mixed edge.
Architecture parameters are jointly trained with the weight parameters in an alternating manner.
However, this optimization process tends to produce a fast-converging architecture, which is not always the optimal solution in terms of accuracy \cite{DBLP:conf/iclr/Shu0C20}.

We propose a new method called minimum stable rank differentiable architecture search (MSR-DARTS) to solve this problem.
With this method, the optimization of the learnable architecture parameters is replaced with the selection process using a stable rank criterion; thus, only weight parameters of neural networks are trained during the architecture search, which enables us to avoid selecting a fast-converging architecture but appropriately find one with a better generalization error.
Our operation set includes only limited convolutional operators ({\it e.g.}, separable convolution and dilated convolution with different kernel sizes), in which each convolutional operator is regarded as a matrix.
We use the stable rank (numerical rank) of each convolution to derive a discrete architecture.
Specifically, in each mixed edge, the operator that has the lowest stable rank is selected.
This architecture search based on stable rank is reasonable since the low-rankness of a matrix indicates the high generalization ability of neural networks.
Several studies \cite{DBLP:conf/icml/Arora0NZ18,DBLP:conf/iclr/SuzukiAN20} reported that a neural network with lower stable rank operators has higher generalization ability, where a stable rank is often used instead of a rank because the former properly captures the statistical degrees of freedom by ignoring negligibly tiny singular values.
More precisely, when an input is noisy, a low stable rank convolution conveys the input information only and attenuates noise. 
Thus, a network with low stable rank has better generalization ability and is robust against noisy input.
In summary, we train an over-parameterized network with fixed uniform architecture parameters; thus, only the weights of the network are optimized.
The discrete architecture is then derived using the stable rank of each operator by treating a convolution as a matrix with MSR-DARTS to yield a well generalized architecture.

We conducted experiments to evaluate MSR-DARTS that involved the CIFAR-10 and ImageNet datasets.
MSR-DARTS achieved an error rate of $2.54\%$ with $4.0$M parameters, which is a lower test error rate than with another DARTS-based method, {\it e.g.}, PC-DARTS \cite{xu2019pc}, and is competitive with Fair DARTS \cite{DBLP:journals/corr/abs-1911-12126}.
MSR-DARTS achieved a top-1 error rate of $23.9\%$ on ImageNet.
To the best of our knowledge, it is a state-of-the-art DARTS-based method that uses CIFAR-10 dataset for architecture search.
The search process takes only $0.3$ GPU-days.
The code is available at \url{https://github.com/mtaecchhi/msrdarts.git}.

\section{Related Work}
\subsection{Neural Architecture Search}
\label{related:NAS}
There has been growing interest in NAS since Zoph and Quoc \cite{DBLP:conf/iclr/ZophL17} proposed its algorithm.
In the early years, EAs and RL were used to optimize network architectures.
The algorithm using RL trains a recurrent neural network meta-controller to guide the search process and  gradually sample a better architecture.
Zoph \etal and Pham \etal \cite{zoph2018learning,DBLP:conf/icml/PhamGZLD18} first optimized the structure of a small component in an entire network, namely a cell, instead of the entire network structure, then constructed the entire network by stacking the optimized cells.
This two-step process reduces search cost.
Liu \etal, Tang \etal and Real \etal \cite{DBLP:conf/iclr/LiuSVFK18,DBLP:conf/icml/TangGD17,DBLP:conf/aaai/RealAHL19} used EAs, which mutate the architecture topologies and evolve towards better performances.
DARTS introduces a differentiable NAS pipeline, which relaxes the search space to be continuous and directly uses gradient-based optimization.
Many studies \cite{DBLP:conf/iclr/XieZLL19,chen2019progressive,xu2019pc,liang2019darts+} followed this approach and achieved remarkable performance with improved efficiency.

\subsection{DARTS}
\label{method:darts}
\begin{figure*}[tp]
    \begin{tabular}{ccc}
        \includegraphics[bb=0 0 500 429, width=.65\columnwidth]{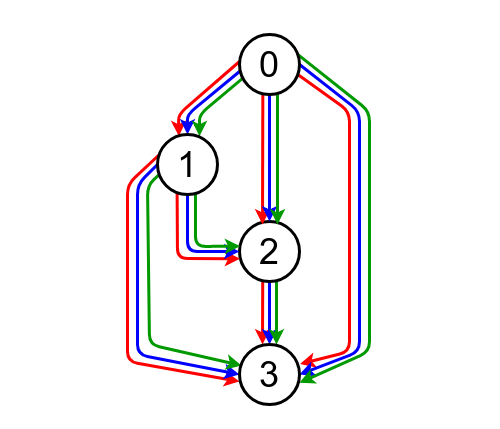} &
        \includegraphics[bb=0 0 700 600, width=.65\columnwidth]{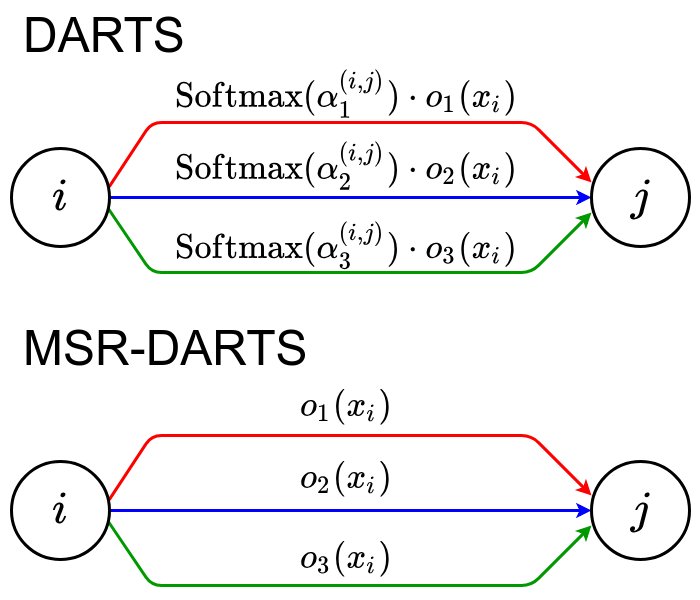} & 
        \includegraphics[bb=0 0 600 514, width=.65\columnwidth]{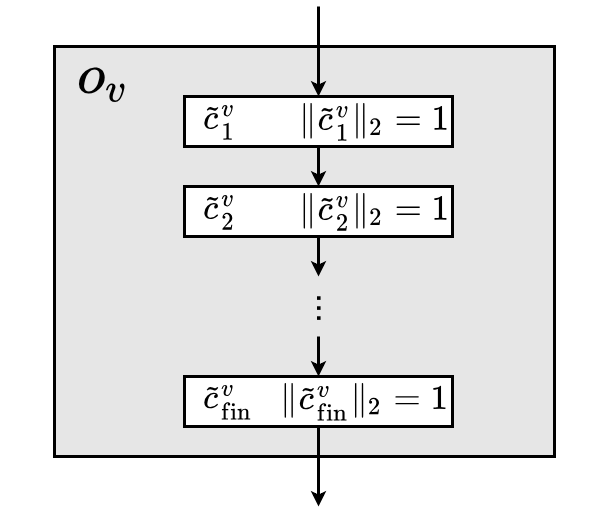}
    \end{tabular}
    \caption{Left: example of DAG with 4 nodes. Nodes and information flows are denoted with circles and arrows, respectively. Middle: difference in information flow between nodes between DARTS (top) and MSR-DARTS (bottom). Each figure indicates that number of candidate operators is $3$. Right: setting spectral norm of each convolution to constant value in search stage.}
    \label{fig:dag_edge_ops} 
\end{figure*}

In this study, similar to previous ones \cite{DBLP:conf/iclr/XieZLL19,chen2019progressive,xu2019pc}, we used DARTS as a baseline framework.
DARTS stacks $L$ cells, each of which is represented as a DAG of an ordered sequence of $N$ nodes, $\{x_0, x_1, \dots, x_{N-1}\}$, where each node $x_i$ is a feature map and each edge $(i,j)\ (i<j)$ denotes an information flow from node $i$ to node $j$.
A set of $K$ candidate operators is denoted as $\mathcal{O} = \{o_0, o_1, \dots, o_{K-1}\}$, in which an element $o_v$ that includes a learnable parameter $w^{(i,j)}_v$ is the $v$-th candidate operator defined in advance ({\it e.g.}, separable convolution and dilated convolution).
Figure \ref{fig:dag_edge_ops} left illustrates an example of a DAG with $4$ nodes.
An information flow between nodes is a mixed edge, which includes all candidate operators.
An architecture parameter $\alpha^{(i,j)}_v$, which indicates how suitable $o_v$ is for mixed edge $(i,j)$, is introduced as a weight for $o_v$.
The goal is to optimize $\alpha^{(i,j)}_v\ (v=0,\dots,K-1)$. 
For each $i < j$, the information flow from node $i$ to node $j$, illustrated in the upper middle of Figure \ref{fig:dag_edge_ops}, is computed as
\begin{equation}
    \label{mixed_op_darts}
    f_{i,j}(x_i) = \sum^{K-1}_{v=0} \frac{\exp(\alpha_v^{(i,j)})}{\sum^{K-1}_{v'=0} \exp(\alpha_{v'}^{(i,j)})} \cdot o_v(x_i),
\end{equation}
where $o_v(x_i)$ is the result of applying the input $x_i$ to $o_v$. 
An output of node $j$ is a sum over all information flows from its predecessors, that is, $x_j = \sum_{i<j}f_{i,j}(x_i)$.
The first two nodes, $x_0$ and $x_1$, are input nodes to a cell.
The output of all the cells equals that of the final node $x_{N-1}$, which is defined as the concatenation of all the intermediate cells, {\it i.e.}, ${\rm concat}(x_2,x_3,\dots,x_{N-2})$.

There are two types of cells introduced in DARTS.
One is a normal cell, which maintains spatial resolution, and the other is a reduction cell, which reduces the spatial resolution of feature maps.
Note that while DARTS shares the architecture topology among all normal cells, the same is true for reduction cells because it optimizes architecture parameters $\alpha_{{\rm normal}}$ and $\alpha_{{\rm reduce}}$, where $\alpha_{{\rm normal}}$ is shared by all the normal cells and $\alpha_{{\rm reduce}}$ is shared by all the reduction cells.

DARTS has two stages.
The first is the search stage, which trains the over-parameterized network consisting of mixed edges in each of which all possible operators are included, and derives a promising discrete architecture in accordance with the architecture parameter $\alpha_v^{(i,j)}$ (see Subsec. \ref{method:deriving_discrete_architecture} for more details).
The second stage is an evaluation stage in which the derived architecture from full-scratch is trained.

\subsection{Generalization Error Analysis}
\label{related:generalization}
A deep network generalizes well even when it has a larger amount of parameters than the sample size.
Generalization bounds represent the upper bound of the generalization error.
We can guarantee that a model with small training error can have a high generalization performance if the generalization error (difference between training error and expected error) can be evaluated properly and has a small upper bound.
There are several metrics to represent generalization error bounds such as VC-dimension \cite{Vapnik1998,pmlr-v65-harvey17a}, PAC-Bayes theory \cite{neyshabur2017pac,dziugaite2017computing}, norm based analysis \cite{bartlett2002rademacher,NIPS2015_5797,golowich2018size}, and the compression based approach \cite{DBLP:conf/icml/Arora0NZ18,DBLP:conf/iclr/BaykalLGFR19,DBLP:conf/iclr/SuzukiAN20}.
The compression based approach has recently attracted much attention because it gives tighter generalization error bounds for deep neural networks.
For example, Arora \etal \cite{DBLP:conf/icml/Arora0NZ18} used the low rank property of weight matrices to compress neural networks based on the fact that a matrix with a lower stable rank is more robust to noise thus generalizes well.
Suzuki \etal \cite{DBLP:conf/iclr/SuzukiAN20} experimentally confirmed that most networks have near low rank weight matrices after training, where a near low rank matrix is defined as a matrix in which a small number of singular values are significantly large while the other singular values are close to zero.
This near low rank property is used to derive generalization bounds.

\section{Method}
\subsection{MSR-DARTS}
\label{method:MSR-DARTS}
While DARTS and many related methods \cite{DBLP:conf/iclr/XieZLL19,chen2019progressive,xu2019pc} search for architectures by jointly training the architecture parameters $\alpha$ and weights of neural networks $w$ ({\it e.g.}, the parameters of the convolutional operator) in an alternating manner ($\alpha^{(i,j)}_v$ is denoted as $\alpha$ and $w^{(i,j)}_v$ is denoted as $w$ for simplicity), Shu \etal \cite{DBLP:conf/iclr/Shu0C20} reported that these methods tend to lead to a fast converged architecture during the search stage rather than well generalized models. 
This is because $\alpha$ is updated on the basis of $w$, which is not fully converged rather than well trained $w$, which is harmful especially in the early epochs.
This means that there is room to search for a model with lower error rate in the search space. 
MSR-DARTS addresses this issue. 
It fixes $\alpha$ to $1$, hence optimizes only $w$. 
In the search stage, instead of Eq. (\ref{mixed_op_darts}), the information flow from node $i$ to node $j$ with MSR-DARTS illustrated in Figure \ref{fig:dag_edge_ops} middle bottom, is defined as 
\begin{equation}
    \label{mixed_op_wgdarts}
    f_{i,j}(x_i) = \sum_{o \in \mathcal{O}} o(x_i).
\end{equation}
The other process is the same as those defined in Subsec. \ref{method:darts}.

With MSR-DARTS, we assume that each candidate operator $o_v \in \mathcal{O}$ consists of several convolutional layers.
Let $c^v_p$ and $c^v_{{\rm fin}}$ be the $p$-th and last convolutional layers in $o_v$, respectively.
Note that the notation of node $(i,j)$ is omitted for simplicity in $c^v_p$ and $c^v_{{\rm fin}}$.
Regarding convolutional computation as matrix calculation \cite{DBLP:conf/iclr/SedghiGL19}, the stable rank of convolution $c$ is then denoted as $R(c)=\frac{\|c\|^2_F}{\|c\|^2_2}$, where $\|\cdot\|_F$ denotes the Frobenius norm and $\|\cdot\|_2$ denotes the spectral norm.
A stable rank is often used as a surrogate for a rank \cite{DBLP:conf/icml/Arora0NZ18}.

Then we select the best operator from $\mathcal{O}$ for each edge after training of an over-parameterized network in the search stage.
We use the relation between network generalization ability and singular values proposed by Arora \etal \cite{DBLP:conf/icml/Arora0NZ18}.
They reported that a well-generalized network consists of noise-robust convolutions that have low stable ranks.
Noise sensitivity $\psi_{\mathcal{N}}(c,x)$ is defined as 
\begin{equation}
    \label{noise_sensitivity}
    \psi_{\mathcal{N}}(c,x) = \mathbb{E}_{\eta \in \mathcal{N}}\left[\frac{\|c(x+\eta \|x\|) - c(x)\|^2}{\|c(x)\|^2}\right],
\end{equation}
where $c$ is a mapping from real-valued vectors to real-valued vectors ({\it e.g.}, convolutional computation represented by a matrix), $x$ is a vector to be multiplied ({\it i.e.}, input for a convolutional layer), $\mathcal{N}$ is a noise distribution, $\mathbb{E}$ is an expectation value, and $\| \cdot \|$ is the Euclidean norm.
Low noise sensitivity indicates that the convolution matrix has a near low rank property ({\it i.e.}, a low stable rank) because the {\it signal} $x$ is correctly carried whereas {\it noise} $\eta$ is attenuated.
Note that $\psi_{\mathcal{N}}(c,x)$ is at least its stable rank when noise is generated from standard normal distribution, {\it i.e.}, $\eta \sim \mathcal{N}(0,I)$.
For more details, refer to the paper by Arora \etal \cite{DBLP:conf/icml/Arora0NZ18}.

We assume that an operator that generalizes better has lower noise sensitivity, {\it i.e.}, a lower stable rank (experimentally shown in Appendix \ref{appendix:compare_max_stable_rank}).
We further assume that the last convolution $c^v_{{\rm fin}}$ is the most relevant to the output of $o_v$, thus use the stable rank of $c^v_{{\rm fin}}$ ({\it i.e.}, $R(c^v_{{\rm fin}}))$ only.
The operator that has the lowest $R(c^v_{{\rm fin}})$ is selected to yield the discrete architecture.

\subsection{Search Space}
\label{method:search_space}
DARTS defines an operator set with eight operators.
That is, 3$\times$3 and 5$\times$5 separable convolutions, 3$\times$3 and 5$\times$5 dilated separable convolutions, 3$\times$3 max pooling, 3$\times$3 average pooling, identity, and {\it zero}.
Our operator set is the subset of the search space of DARTS, {\it i.e.}, 3$\times$3 and 5$\times$5 separable convolutions and 3$\times$3 and 5$\times$5 dilated separable convolutions.
As described in Subsec. \ref{method:MSR-DARTS}, we assume that all candidate operators $o_v \in \mathcal{O}$ consist of limited convolutional layers, where 3$\times$3 max pooling, 3$\times$3 average pooling, identity, and {\it zero}, which are not convolutional calculations, are excluded.
As with DARTS, we use the ReLU-Conv-BN order for convolutional operators, in which each separable convolution is always applied twice.

\subsection{Setting Spectral Norm of Convolution in Search Stage}
\label{method:setting_spectral_norm}
We assume that with MSR-DARTS, all operators in a search space are trained {\it correctly} in the search stage.
However, previous studies \cite{DBLP:conf/iclr/SuzukiAN20,DBLP:conf/nips/GunasekarLSS18,DBLP:conf/iclr/JiT19} reported that deep learning tends to produce a simpler model than its full expression ability when we use regularization such as $L_1$ regularization. 
More precisely, it has been experimentally shown that a trained network tends to have near low rank weight matrices, in which only a few singular values are large and others are close to zero.
Therefore, in each mixed edge of an over-parameterized network with MSR-DARTS, some operators can be redundant because their singular values can be all close to zero.
To train each operator stably, we apply the spectral norm adjustment technique introduced by Behrmann \etal \cite{DBLP:conf/icml/BehrmannGCDJ19} to all convolutional operators in candidate operators, that is, 
\begin{equation}
    \forall v, p; \ \|c^v_p\|_2 = C
\end{equation}
where $C$ is a constant value to be set (Figure \ref{fig:dag_edge_ops} right).
We estimate the spectral norm of $c^v_p$ by carrying out power-iteration (see Appendix \ref{appendix:power_iteration} for more details), which yields an under-estimate $\tilde{\sigma}_1(c^v_p) \leq \|c^v_p\|_2$, where $\sigma_q(c^v_p)$ denotes the $q$-th largest singular value of $c^v_p$.
Similar to the above study, using this estimate, each convolution in candidate operator $c^v_p$ is normalized as
\begin{equation}
    \tilde{c}^v_p = c^v_p \cdot \frac{C}{\tilde{\sigma}_1(c^v_p)}.
\end{equation}
Note that spectral norm adjustment of each convolution is conducted before forward propagation.
We used $C=1$ in our experiments.

\subsection{Deriving Discrete Architecture}
\label{method:deriving_discrete_architecture}
After training of the over-parameterized network, a discrete architecture is derived by selecting the topology and operator for each intermediate node.
With DARTS and its derivations, the topology is selected by retaining two of the strongest precedent edges for each intermediate node, where the strength of an edge from node $i$ to node $j$, denoted as $S_{i,j}$, is defined as 
\begin{equation}
    \label{derive_topology}
    S_{i,j} = \max_{o \in \mathcal{O},o \neq zero} \frac{\exp(\alpha_o^{(i,j)})}{\sum_{o' \in \mathcal{O}} \exp(\alpha_{o'}^{(i,j)})}.
\end{equation}
In each edge, the operator with the largest $\alpha_v$ is selected.
However, as described in Subsec. \ref{method:MSR-DARTS}, MSR-DARTS uses a fixed value for $\alpha_v$.
We use the stable rank of the last convolution $R(c^v_{\rm fin})$ in each $o_v$ to determine the topology.
First, for normal and reduction cells, the mixed edge between node $i$ and node $j$ is replaced with the best operator $(o^{\ast}_{\mathcal{T}})_{i,j}$, defined by
\begin{equation}
    \label{eq:best_op}
    (o^{\ast}_{\mathcal{T}})_{i,j}= \argmin_{o_v \in \mathcal{O}} \ \ \overline{R}_{\mathcal{T}}(c^{(i,j)v}_{\rm fin}).
\end{equation}
We explicitly notate the edge indices $i,j$ in convolution $c$, $\mathcal{T}=\{ {\rm normal}, {\rm reduce}\}$ denotes the type of cells and $\overline{R}_{\mathcal{T}}(\cdot)$ denotes the average $R(\cdot)$ of all the cells belonging to cell type $\mathcal{T}$.
Similar to Eq. (\ref{eq:best_op}), the strength of an edge from node $i$ to node $j$ of cell type $\mathcal{T}$ with MSR-DARTS is defined by 
\begin{equation}
    \label{wg_derive_topology}
    (S_{\mathcal{T}})_{i,j} = \max_{o_v \in \mathcal{O}} \left(-\overline{R}_{\mathcal{T}}(c^{(i,j)v}_{\rm fin}) \right)
\end{equation}
We use Eq. (\ref{wg_derive_topology}) instead of Eq. (\ref{derive_topology}) to derive topology.
Note that the operator $o_v \in \mathcal{O}$, which has lower $R(c^v_{\rm fin})$ is considered a better operator in terms of generalization ability (see Subsec. \ref{method:MSR-DARTS}).

As with DARTS, each intermediate node retains the connections from the two strongest precedent nodes, denoted as $i^{\ast}_1$ and $i^{\ast}_2$, with Eq. (\ref{wg_derive_topology}), that is, 
\begin{eqnarray}
    \label{strongest_nodes}
    i^{\ast}_1 = \argmax_{i} \ \ S_{i,j} \nonumber \\ 
    i^{\ast}_2 = \argmax_{i \neq i^{\ast}_1} \ \ S_{i,j}.
\end{eqnarray}
Note that the cell-type notation $\mathcal{T}$ is omitted from Eq. (\ref{strongest_nodes}) for simplicity.

In summary, a discrete architecture is derived from the over-parameterized network after training, where each mixed edge is replaced with the best operator $(o^{\ast}_{\mathcal{T}})_{i,j}$ in accordance with Eq. (\ref{eq:best_op}).
The two strongest precedent connections from node $i^{\ast}_1$ and $i^{\ast}_2$ are then preserved in each node in accordance with Eq. (\ref{wg_derive_topology}).

\section{Experiments}
\label{experiments}
\subsection{Datasets}
\label{experiments:datasets}
Similar to several previous studies \cite{liu2018darts,chen2019progressive,xu2019pc}, we conducted experiments on the well-known image classification datasets CIFAR-10 \cite{Krizhevsky09learningmultiple} and ImageNet \cite{imagenet_cvpr09}.
CIFAR-10 consists of 50K training images and 10K testing images. 
All images have a fixed size of 32$\times$32 and equally distributed over ten classes.
ImageNet was obtained from the Imagenet Large Scale Visual Recognition Challenge 2012 \cite{ILSVRC15} and contains 1K object classes, 1.28M training images, and 50K validation images.
We follow the general setting in which the input image size is $224 \times 224$.

\subsection{CIFAR-10 Toy Experiment}
\label{experiments:toy}
First, we confirmed the effectiveness of the proposed method using MSR through a simple toy experiment using CIFAR-10.
In this experiment, MSR-DARTS and DARTS \cite{liu2018darts} were compared, both under the same experimental conditions.
The results indicate that MSR-DARTS is superior to DARTS.

In the search stage, following DARTS, the architecture was conducted on a network with $L=8$ cells.
Each convolutional cell consists of $N=7$ nodes.
The input nodes $x_0$ and $x_1$ were equal to the output of the last two preceding cells, respectively.
The output node was $x_6$, which is the concatenation of all the intermediate nodes.
Reduction cells were inserted at 1/3 and 2/3 the total depth of the network to reduce the spatial resolution of feature maps.
All other cells were normal cells that maintain spatial resolution. 
Note that while DARTS and related methods \cite{DBLP:conf/iclr/XieZLL19,xu2019pc} share the architecture topology among all normal cells, the same is true for reduction cells, because they optimize architecture parameters $(\alpha_{{\rm normal}}, \alpha_{{\rm reduce}})$, where $\alpha_{{\rm normal}}$ is shared by all the normal cells and $\alpha_{{\rm reduce}}$ is shared by all the reduction cells (see Subsec. \ref{method:darts}).
MSR-DARTS also optimizes two types of cell structures, {\it i.e.}, normal and reduction cell, but each cell is optimized using MSR (see Subsec. \ref{method:MSR-DARTS}) because learnable architecture parameters are not optimized with MSR-DARTS.

\subsubsection{Experimental Settings}
\label{toy:experimental_setting}
We split the CIFAR-10 training data in a ratio of four to one.
The former was used to train the over-parameterized network weights while the latter was used to calculate the loss.
The over-parameterized network was trained for 50 epochs, where the batch size was determined to fit into a single GPU.
The initial number of channels was set to 16. 
We used momentum stochastic gradient descent (SGD) to optimize the weights, with an initial learning rate of 0.025 (annealed down following a cosine schedule), momentum of 0.9, and weight decay of $3\times10^{-4}$.
We used search space $\mathcal{O}_{\rm MSR} \in $ \{``3x3 separable conv'', ``5x5 separable conv'', ``3x3 dilated conv'', ``5x5 dilated conv''\}.
Note that we used search space $\mathcal{O}_{\rm MSR}$ because with MSR-DARTS assumed that each candidate operator in the search space is composed of several convolutional layers (see Subsec \ref{method:MSR-DARTS}); therefore, we omitted the operators that do not meet this assumption, {\it i.e.}, ``zero'', ``identity'', ``3x3 max pooling'', ``3x3 average pooling'' from the search space.
We used single the TITAN RTX GPU, and the architecture search took less than $0.3$ days.

In the evaluation stage of the toy experiment, the network was composed of $8$ cells (6 normal cells and 2 reduction cells) instead of $20$ cells (18 normal cells and 2 reduction cells), the number of which is widely used to compare the performance of DARTS-based methods.
The other settings follow those of DARTS.
The network was trained for 600 epochs, with a batch size 96.
The initial number of channels was 36, the SGD optimizer with an initial learning rate of $0.025$ (annealed down to zero following a cosine schedule), a momentum of $0.9$, a weight decay of $3 \times 10^{-4}$ and gradient clipping at $5$.
Cutout \cite{devries2017improved}, path dropout of probability $0.3$, and auxiliary towers with weight $0.4$ were used to enhance accuracy.

\subsubsection{Results}
\label{toy:results}
\begin{figure}[tp]
    \begin{center}
        \includegraphics[bb=0 0 336 212, width=.75\columnwidth]{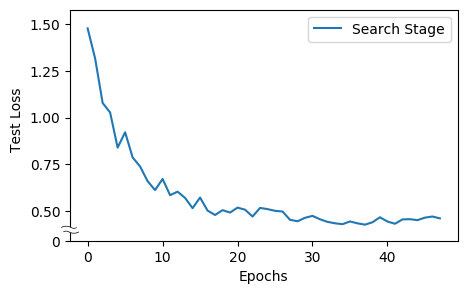}
        \caption{Validation-loss transition of over-parameterized network in search stage. 50,000 training images were split into ratio of 4:1, former was used to optimize network parameters, latter was used to validate network.}
        \label{fig:search_layer8_test_loss}        
    \end{center}
\end{figure}
\begin{figure}[tp]
    \includegraphics[bb=0 0 2024 657, width=1.0\columnwidth]{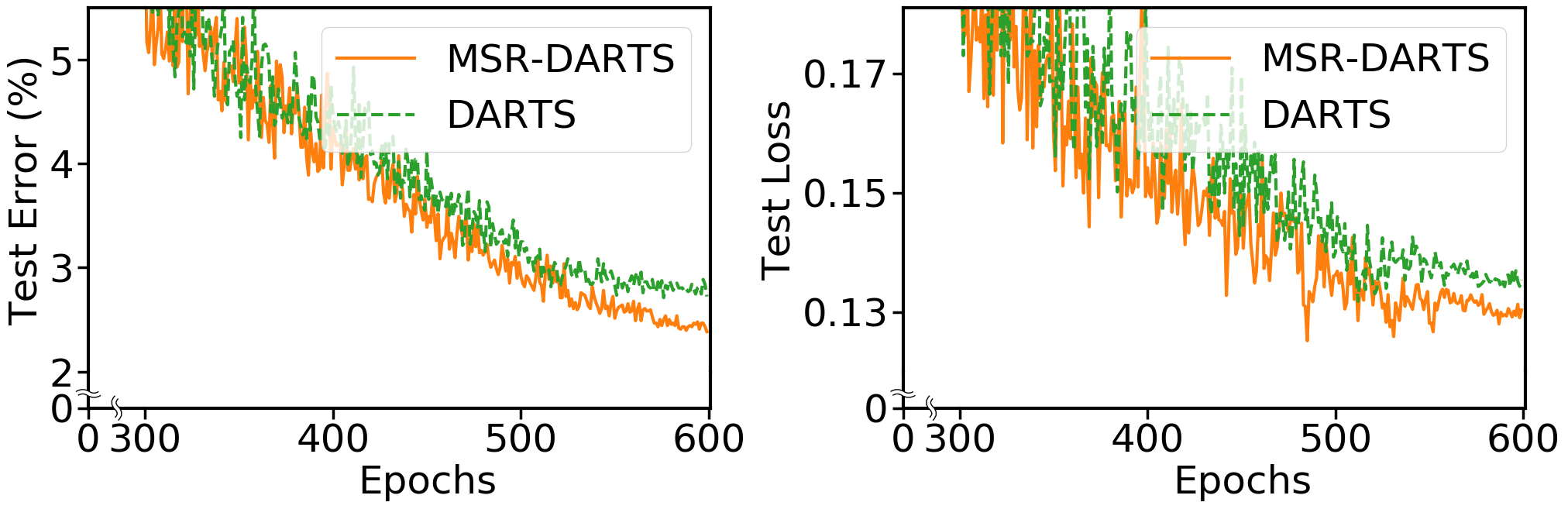}
    \caption{Test-error transition (left) and test-loss transition (right) of networks generated with MSR-DARTS and DARTS in evaluation stage of toy experiment (after 300 epochs were plotted). Test loss is average of cross entropy loss over all test sets. Orange solid lines represent MSR-DARTS and green dotted lines represent DARTS. }
    \label{fig:compare_msvdarts_darts_loss_acc} 
\end{figure}
The validation loss of the over-parameterized network in the search stage is illustrated in Figure \ref{fig:search_layer8_test_loss}. 
In the search stage, the validation loss decreased from the beginning of training and took a minimum value around 35 epochs.
However, it began to increase little by little after 35 epochs.
Since MSR-DARTS aims to determine a model structure with high generalization performance, the architecture generated before the increase in the validation loss was used for architecture evaluation.
Specifically, we used the architecture generated in epoch 38.

\begin{table}[tp]
    \begin{center}
        \caption{Comparison with DARTS in toy experiment. Search space $\mathcal{O}_{\rm MSR}$ was used. Each architecture was searched with $8$ cells in search stage and evaluated $8$ cells in evaluation stage. Note that column {\bf S.C} is search cost (GPU-days). }
        \label{table:compare_cifar10_8layers}
        \begin{tabular}{lrrr} \bhline{1.5pt}
            {\bf Architecture} & \begin{tabular}{r} {\bf Test Err.}\\{\bf (\%)} \end{tabular}  & {\bf S.C.} & \begin{tabular}{r} {\bf Params}\\{\bf (M)} \end{tabular} \\ \hline
            DARTS & $3.30$ & $1.0$ & 1.65 \\
            MSR-DARTS & ${\bf 2.84}$ & ${\bf 0.3}$ & ${\bf 1.63}$ \\ \bhline{1.5pt} 
        \end{tabular}
    \end{center}
\end{table}
Figure \ref{fig:compare_msvdarts_darts_loss_acc} left and right indicate the test accuracy and test loss, respectively.
Each value is plotted after 300 epochs where the difference can be observed.
The architecture generated with MSR-DARTS exhibited lower test error and higher test accuracy than those with DARTS, which indicates MSR-DARTS outperforms DARTS when the layer depth in the search stage is the same as that in the evaluation stage.
Table \ref{table:compare_cifar10_8layers} summarizes the results of the toy experiment.
MSR-DARTS achieved a test error rate of $2.84\%$, which is $0.46\%$ higher than that of DARTS.
MSR-DARTS only requires $0.3$ days with a GPU to search for the optimal architecture, which is 3 times faster than with DARTS.
In terms of comparing the number of parameters of each architecture, there was almost no difference between the models since the same search space was used.
Strictly speaking, the number of parameters of MSR-DARTS is $1.63$M, which is smaller than that of DARTS.

To conclude the toy experiment, we set the same experimental settings such as search space and hyper-parameters for both methods and confirmed that MSR-DARTS yields better results than DARTS.

\subsection{Results on CIFAR-10}
\label{experiments:cifar10}
\begin{table*}[tph]
    \begin{center}
        \caption{Comparison with state-of-the-art image classifiers on CIFAR-10. Cutout is denoted as c/o.}
        \label{table:compare_cifar10}
        \begin{tabular}{lrrrl} \bhline{1.5pt}
            {\bf Architecture} & \begin{tabular}{r} {\bf Test Err.}\\{\bf (\%)} \end{tabular} & \begin{tabular}{r} {\bf Params}\\{\bf (M)} \end{tabular} & \begin{tabular}{r} {\bf Search Cost}\\{\bf (GPU-days)} \end{tabular} & {\bf Search Method}  \\ \hline
            DenseNet-BC \cite{Huang_2017_CVPR} & $3.46$ & $25.6$ & - & manual \\ \hline
            NASNet-A + c/o \cite{zoph2018learning} & $2.65$ & $3.3$ & $1800$ & RL \\
            AmoebaNet-B + c/o \cite{DBLP:conf/aaai/RealAHL19} & $2.55$ & $2.8$ & $3150$ & evolution \\
            Hierarchical Evo \cite{DBLP:conf/iclr/LiuSVFK18} & $3.75$ & $15.7$ & $300$ & evolution \\
            PNAS \cite{Liu_2018_ECCV} & $3.41$ & 3.2 & $225$ & SMBO \\
            ENAS + c/o \cite{DBLP:conf/icml/PhamGZLD18} & $2.89$ & $4.6$ & $0.5$ & RL \\ \hline
            DARTS (1st order) + c/o \cite{liu2018darts} & $3.00$ & $3.3$ & $0.4$ & gradient-based \\
            DARTS (2nd order) + c/o \cite{liu2018darts} & $2.76$ & $3.3$ & $1.0$ & gradient-based \\
            SNAS (moderate) + c/o \cite{DBLP:conf/iclr/XieZLL19} & $2.85$ & $2.8$ & $1.5$ & gradient-based \\
            ProxylessNAS + c/o \cite{DBLP:conf/iclr/CaiZH19} & $2.08$ & - & $4.0$ & gradient-based \\
            DARTS+ + c/o \cite{liang2019darts+} & $2.20$ & $4.3$ & $0.6$ & gradient-based \\
            P-DARTS + c/o \cite{chen2019progressive} & $2.50$ & $3.4$ & $0.3$ & gradient-based \\
            PC-DARTS + c/o \cite{xu2019pc} & $2.57$ & $3.6$ & $0.1$ & gradient-based \\ 
            Fair DARTS + c/o \cite{DBLP:journals/corr/abs-1911-12126} & $2.54$ & $2.8$ & $0.4$ & gradient-based \\
            PR-DARTS + c/o \cite{zhou2020theory} & $2.32$ & $3.4$ & $0.17$ & gradient-based \\ \hline
            MSR-DARTS + c/o (ours) & $2.54$ & $4.0$ & $0.3$ & Stable Rank \\ \bhline{1.5pt}
        \end{tabular}            
    \end{center}
\end{table*}
We evaluated the MSR-DARTS with $20$ cells (18 normal cells and 2 reduction cells) in the evaluation stage to compare other state-of-the-art methods.
We used the same cell structure optimized in the toy experiment.
The rest of the experimental settings were exactly the same as with DARTS.
Table \ref{table:compare_cifar10} compares the image-classification performance of MSR-DARTS with those of the other methods.
MSR-DARTS achieved $2.54\%$ test error, which is better than 2nd order DARTS.
It outperformed PC-DARTS \cite{xu2019pc} and was competitive with Fair DARTS \cite{DBLP:journals/corr/abs-1911-12126} in terms of test-error rate.
Our architecture search finished in $0.3$ GPU-days, which is faster than 2nd order DARTS ($1.0$ GPU-day) and Fair DARTS ($0.4$ GPU-days).
Note that MSR-DARTS generated the optimal model with $4.0$M parameters, which is a slightly larger number of parameters compared with the models of the other methods.
This is because of search space $\mathcal{O}_{\rm MSR}$, which includes the convolutional operator only.
This means that the operator with few parameters, such as pooling or identity (skip connection), is not selected.

\subsection{Results on ImageNet}
\label{experiments:imagenet}
\begin{table*}[tp]
    \begin{center}
        \caption{Comparison with state-of-the-art image classifiers on ImageNet. $\dagger$ denotes architecture was searched for on ImageNet directly.}
        \label{table:compare_imagenet}
        \begin{tabular}{lrrrrrl} \bhline{1.5pt}
            {\bf Architecture} & \multicolumn{2}{r}{{\bf Test Err.(\%)}} & {\bf Params} &$\times+$ &{\bf Search Cost} & {\bf Search Method}\\ \cline{2-3}
            {} & {\bf top-1} & {\bf top-5} & {\bf (M)} & {\bf (M)} & {\bf GPU-days} & {} \\\hline 
            Inception-v1 \cite{Szegedy_2015_CVPR} & $30.2$ & $10.1$ & $6.6$ & $1448$ & - & manual \\
            MobileNet \cite{DBLP:journals/corr/HowardZCKWWAA17} & $29.4$ & $10.5$ & $4.2$ & $569$ & - & manual \\
            ShuffleNet $2\times$ (v1) \cite{DBLP:conf/cvpr/ZhangZLS18} & $26.4$ & $10.2$ & $~5$ & $524$ & - & manual \\
            ShuffleNet $2\times$ (v2) \cite{Ma_2018_ECCV} & $25.1$ & - & $~5$ & $591$ & - & manual \\ \hline

            NASNet-A \cite{zoph2018learning} & $26.0$ & $8.4$ & $5.3$ & $564$ & $1800$ & RL \\ 
            AmoebaNet-C \cite{DBLP:conf/aaai/RealAHL19} & $24.3$ & $7.6$ & $6.4$ & $570$ & $3150$ & evolution \\ 
            PNAS  \cite{Liu_2018_ECCV} & $25.8$ & $8.1$ & $5.1$ & $588$ & $225$ & SMBO \\ 
            MnasNet-92 \cite{Tan_2019_CVPR} & $25.2$ & $8.0$ & $4.4$ & $388$ & - & RL \\ \hline

            DARTS (2nd order) \cite{liu2018darts} & $26.7$ & $8.7$ & $4.7$ & $574$ & $4.0$ & gradient-based \\
            SNAS (mild) \cite{DBLP:conf/iclr/XieZLL19} & $27.3$ & $9.2$ & $4.3$ & $522$ & $1.5$ & gradient-based \\
            ProxylessNAS(GPU) \footnotemark[2]  \cite{DBLP:conf/iclr/CaiZH19} & $24.9$ & $7.5$ & $7.1$ & $465$ & $8.3$ & gradient-based \\
            DARTS+(CIFAR-100)\cite{liang2019darts+} & $23.7$ & $7.2$ & $5.1$ & $591$ & $0.2$ & gradient-based \\
            DARTS+(ImageNet) \footnotemark[2]  \cite{liang2019darts+}& $23.9$ & $7.4$ & $5.1$ & $582$ & $6.8$ & gradient-based \\
            P-DARTS(CIFAR-10) \cite{chen2019progressive} & $24.4$ & $7.4$ & $4.9$ & $557$ & $0.3$ & gradient-based \\
            P-DARTS(CIFAR-100) \cite{chen2019progressive} & $24.7$ & $7.5$ & $5.1$ & $577$ & $0.3$ & gradient-based \\
            PC-DARTS(CIFAR-10) \cite{xu2019pc} & $25.1$ & $7.8$ & $5.3$ & $586$ & $0.1$ & gradient-based \\
            PC-DARTS(ImageNet) \footnotemark[2] \cite{xu2019pc} & $24.2$ & $7.3$ & $5.3$ & $597$ & $3.8$ & gradient-based \\ 
            Fair DARTS \cite{DBLP:journals/corr/abs-1911-12126} & $24.9$ & $7.5$ & $4.8$ & $541$ & $0.4$ & gradient-based \\
            PR-DARTS \cite{zhou2020theory} & $24.1$ & $7.3$ & $4.98$ & $541$ & $0.17$ & gradient-based \\ \hline
            MSR-DARTS(CIFAR-10) (ours) & $23.9$ & $7.3$ & $5.6$ & $632$ & $0.3$ & Stable Rank \\ \bhline{1.5pt}
        \end{tabular}
        \footnotetext[2]{This architecture was searched for on ImageNet directly.}
    \end{center}
\end{table*}
Following DARTS and its derivations, we evaluated the architecture with $L=14$ cells in the architecture evaluation stage.
We used the normal and reduction cells, which were optimized with CIFAR-10 (see Subsec. \ref{experiments:toy}).
The experimental settings mostly followed those of DARTS.
The network was trained $250$ epochs with a batch size $128$.
The initial number of channels was set to be $48$.
We used momentum SGD to optimize the weights, with an initial learning rate $0.1$ (annealed down following a cosine schedule), momentum of $0.9$, and weight decay of $3 \times 10^{-4}$.
For additional enhancements, label smoothing and an auxiliary loss tower were used during the training.
We used $32$ TITAN V100 GPUs, applied learning warm-up for the first $10$ epochs (increase linearly with each batch), and used Horovod for distributed training. 

The results are summarized in Table \ref{table:compare_imagenet}.
The architecture found with MSR-DARTS using CIFAR-10 reported a top-1 error rate of $23.9\%$ and top-5 error rate of $7.3\%$, which outperforms the top-1 error rate of $26.7\%$ and top-5 error rate of $8.7\%$ reported with DARTS.
In the comparison with other DARTS-based methods using CIFAR-10 for architecture search, MSR-DARTS had $0.5\%$ higher top-1 accuracy than P-DARTS, $1.2\%$ higher than PC-DARTS, $1.0\%$ higher than Fair DARTS, and  $0.2\%$ higher than PR-DARTS. 
It was also competitive with $23.9\%$ of DARTS+, which searches for an optimal architecture with CIFAR-100.
Regarding search cost, MSR-DARTS required only $0.3$ GPU-days, which is 3 times faster than the 2nd order DARTS and slightly faster or competitive with Fair DARTS and P-DARTS.
Due to the fact that the search space includes only limited types of convolutional operators (see Subsec. \ref{method:search_space}), MSR-DARTS had a relatively large number of parameters and multiply-add operations.

\subsection{Qualitative Evaluation}
\label{experiments:qualitative_evaluation}
\begin{figure*}[tp]
  \begin{center}
    \subfigure[Normal cell found with DARTS.]{\includegraphics[bb=0 0 829 367, width=.9\columnwidth]{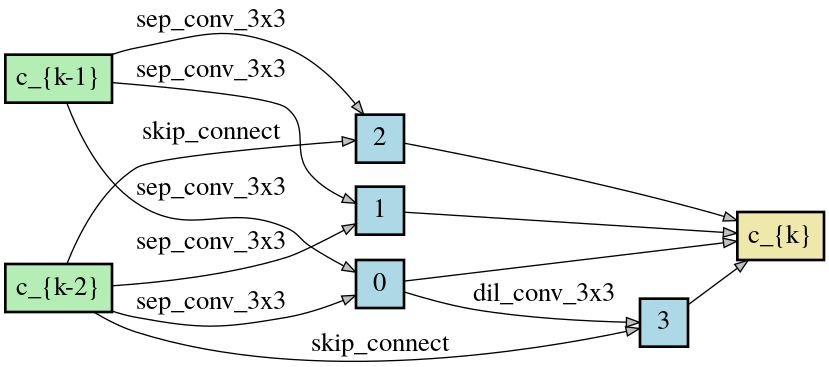}}
    \label{fig:darts_normal_cell}
    \subfigure[Reduction cell found with DARTS.]{\includegraphics[bb=0 0 851 336, width=.9\columnwidth]{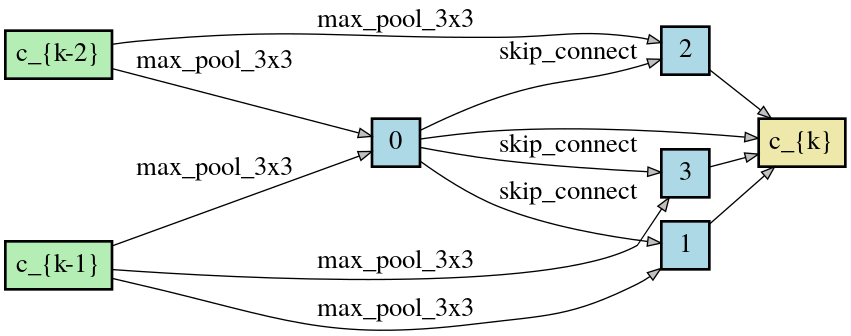}}
    \label{fig:darts_reduction_cell}    
    \subfigure[Normal cell found with MSR-DARTS.]{\includegraphics[bb=0 0 1088 288, width=.9\columnwidth]{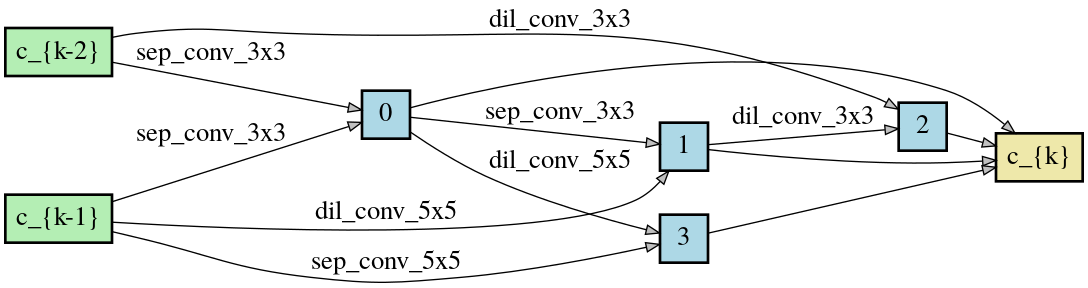}}
    \label{fig:msr_normal_cell}
    \subfigure[Reduction cell found with MSR-DARTS.]{\includegraphics[bb=0 0 1100 500, width=.9\columnwidth]{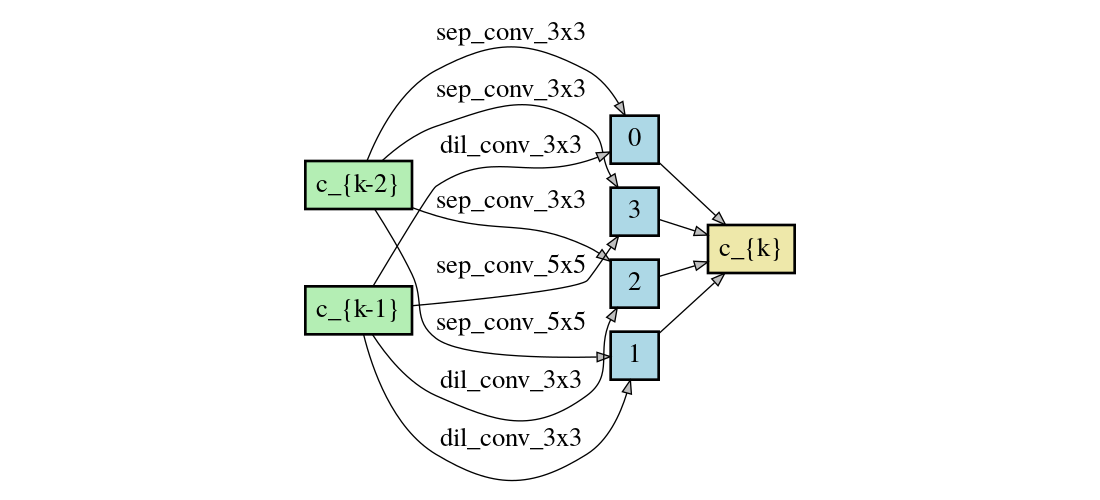}}
    \label{fig:msr_reduction_cell}
  \end{center}
  \caption{Optimized cells found with DARTS and MSR-DARTS.}
  \label{fig:cell_architectures}
\end{figure*}
In this subsection, we compare the optimized architectures between DARTS and MSR-DARTS qualitatively.
Figure \ref{fig:cell_architectures} illustrates the cell structures found with DARTS((a) normal cell, (b) reduction cell) and with MSR-DARTS ((c) normal cell, (d) reduction cell).

As described in a previous paper \cite{DBLP:conf/iclr/Shu0C20}, the depth of a cell structure is defined as the number of connections along the longest path from input node to output node.
The width of a cell structure is defined as the total width of intermediate nodes that are connected to the input nodes.
In particular, the normal cell found with DARTS (Figure \ref{fig:cell_architectures} (a)) had a depth of $3$ and width of $3.5$.

The normal cell found by MSR-DARTS, however, had a depth of $4$ and width of $2.5$, a deeper and narrower structure.
As discussed in the above paper \cite{DBLP:conf/iclr/Shu0C20}, DARTS tends to select shallow and wide cells because is optimizes two parameter sets (weight $w$ and architecture parameter $\alpha$) in an alternating manner.
Each candidate architecture is not evaluated based on the generalization performance at convergence during architecture search; thus architectures with faster convergence rates tend to be selected.
In contrast, the architecture parameters are fixed and only $w$ are optimized with MSR-DARTS.
Thus, MSR-DARTS can evaluate each architecture more correctly at convergence and yield deeper and narrower cells, which is a well generalized architecture.
Note that the scheme that penalizes shallower cells is used with PR-DARTS \cite{zhou2020theory}.
However, PR-DARTS has many more hyper-parameters than the original DARTS, which are difficult to adjust.
On the other hand, MSR-DARTS does not require any hyper-parameters to deepen a cell structure, making it easy to optimize.
All the intermediate nodes, except the first intermediate node (illustrated as $0$ in Figure \ref{fig:cell_architectures}) in the normal cell found with MSR-DARTS had both a connection from an input node and a connection from another intermediate node.
Each intermediate node can handle both the input features of the cell and well-processed features inside the cell.
This structure may improve of the accuracy.

The reduction cell found with MSR-DARTS had a shallow and wide cell structure in contrast to the normal cell.
The reason MSR-DARTS finds these structures may be that the reduction cell should summarize the information to reduce the spatial dimension, thus requiring more information from the input features than the normal cell.
In addition, it seems that the reduction cell found with MSR-DARTS has small receptive fields ({\it e.g.}, ``3x3 separable convolution'' and ``5x5 separable convolution'') for the output of the before last cell and relatively large receptive fields ({\it e.g.}, ``5x5 separable convolution'' and ``3x3 dilated convolution'') for the output of the before cell.
This enables the efficient aggregation of information by applying a large receptive field to the latest information.

\section{Conclusion}
We proposed MSR-DARTS, which optimizes only the weight parameters of an over-parameterized network.
The discrete-architecture-selection process using the optimized architecture parameters with DARTS is replaced with the process using a stable rank of each convolution.
By using the relation between the generalization ability of a neural network and low-rankness of an operator, MSR-DARTS searches for an architecture that is expected to have low test error by selecting the lowest stable rank operator.
In our evaluation, MSR-DARTS achieved $2.54\%$ test error on CIFAR-10 and $23.9\%$ test error on ImageNet.
The architecture-search process can be done within $0.3$ GPU-days.

\subsection*{Acknowledgements}
This work was supported by JST CREST JPMJCR1687 and NEDO JPNP18002. TS was partially supported by MEXT Kakenhi (15H05707, 18K19793 and 18H03201), and Japan Digital Design.

{\small
\bibliographystyle{ieee_fullname}
\bibliography{refs}

\begin{thebibliography}{10}\itemsep=-1pt

\bibitem{ahn2018fast}
Namhyuk Ahn, Byungkon Kang, and Kyung-Ah Sohn.
\newblock Fast, accurate, and lightweight super-resolution with cascading
  residual network.
\newblock In {\em Proceedings of the European Conference on Computer Vision
  (ECCV)}, pages 252--268, 2018.

\bibitem{DBLP:conf/icml/Arora0NZ18}
Sanjeev Arora, Rong Ge, Behnam Neyshabur, and Yi Zhang.
\newblock Stronger generalization bounds for deep nets via a compression
  approach.
\newblock In Jennifer~G. Dy and Andreas Krause, editors, {\em Proceedings of
  the 35th International Conference on Machine Learning, {ICML} 2018,
  Stockholmsm{\"{a}}ssan, Stockholm, Sweden, July 10-15, 2018}, volume~80 of
  {\em Proceedings of Machine Learning Research}, pages 254--263. {PMLR}, 2018.

\bibitem{bartlett2002rademacher}
Peter~L Bartlett and Shahar Mendelson.
\newblock Rademacher and gaussian complexities: Risk bounds and structural
  results.
\newblock {\em Journal of Machine Learning Research}, 3(Nov):463--482, 2002.

\bibitem{DBLP:conf/iclr/BaykalLGFR19}
Cenk Baykal, Lucas Liebenwein, Igor Gilitschenski, Dan Feldman, and Daniela
  Rus.
\newblock Data-dependent coresets for compressing neural networks with
  applications to generalization bounds.
\newblock In {\em 7th International Conference on Learning Representations,
  {ICLR} 2019, New Orleans, LA, USA, May 6-9, 2019}. OpenReview.net, 2019.

\bibitem{DBLP:conf/icml/BehrmannGCDJ19}
Jens Behrmann, Will Grathwohl, Ricky T.~Q. Chen, David Duvenaud, and
  J{\"{o}}rn{-}Henrik Jacobsen.
\newblock Invertible residual networks.
\newblock In Kamalika Chaudhuri and Ruslan Salakhutdinov, editors, {\em
  Proceedings of the 36th International Conference on Machine Learning, {ICML}
  2019, 9-15 June 2019, Long Beach, California, {USA}}, volume~97 of {\em
  Proceedings of Machine Learning Research}, pages 573--582. {PMLR}, 2019.

\bibitem{DBLP:conf/iclr/CaiZH19}
Han Cai, Ligeng Zhu, and Song Han.
\newblock Proxylessnas: Direct neural architecture search on target task and
  hardware.
\newblock In {\em 7th International Conference on Learning Representations,
  {ICLR} 2019, New Orleans, LA, USA, May 6-9, 2019}. OpenReview.net, 2019.

\bibitem{chen2019progressive}
Xin Chen, Lingxi Xie, Jun Wu, and Qi Tian.
\newblock Progressive differentiable architecture search: Bridging the depth
  gap between search and evaluation.
\newblock In {\em Proceedings of the IEEE International Conference on Computer
  Vision}, pages 1294--1303, 2019.

\bibitem{DBLP:journals/corr/abs-1911-12126}
Xiangxiang Chu, Tianbao Zhou, Bo Zhang, and Jixiang Li.
\newblock Fair {DARTS:} eliminating unfair advantages in differentiable
  architecture search.
\newblock {\em CoRR}, abs/1911.12126, 2019.

\bibitem{imagenet_cvpr09}
J. Deng, W. Dong, R. Socher, L.-J. Li, K. Li, and L. Fei-Fei.
\newblock {ImageNet: A Large-Scale Hierarchical Image Database}.
\newblock In {\em CVPR09}, 2009.

\bibitem{devries2017improved}
Terrance DeVries and Graham~W Taylor.
\newblock Improved regularization of convolutional neural networks with cutout.
\newblock {\em arXiv preprint arXiv:1708.04552}, 2017.

\bibitem{dziugaite2017computing}
Gintare~Karolina Dziugaite and Daniel~M Roy.
\newblock Computing nonvacuous generalization bounds for deep (stochastic)
  neural networks with many more parameters than training data.
\newblock {\em arXiv preprint arXiv:1703.11008}, 2017.

\bibitem{golowich2018size}
Noah Golowich, Alexander Rakhlin, and Ohad Shamir.
\newblock Size-independent sample complexity of neural networks.
\newblock In {\em Conference On Learning Theory}, pages 297--299, 2018.

\bibitem{DBLP:conf/nips/GunasekarLSS18}
Suriya Gunasekar, Jason~D. Lee, Daniel Soudry, and Nati Srebro.
\newblock Implicit bias of gradient descent on linear convolutional networks.
\newblock In Samy Bengio, Hanna~M. Wallach, Hugo Larochelle, Kristen Grauman,
  Nicol{\`{o}} Cesa{-}Bianchi, and Roman Garnett, editors, {\em Advances in
  Neural Information Processing Systems 31: Annual Conference on Neural
  Information Processing Systems 2018, NeurIPS 2018, 3-8 December 2018,
  Montr{\'{e}}al, Canada}, pages 9482--9491, 2018.

\bibitem{pmlr-v65-harvey17a}
Nick Harvey, Christopher Liaw, and Abbas Mehrabian.
\newblock Nearly-tight {VC}-dimension bounds for piecewise linear neural
  networks.
\newblock In Satyen Kale and Ohad Shamir, editors, {\em Proceedings of the 2017
  Conference on Learning Theory}, volume~65 of {\em Proceedings of Machine
  Learning Research}, pages 1064--1068, Amsterdam, Netherlands, 07--10 Jul
  2017. PMLR.

\bibitem{DBLP:journals/corr/HowardZCKWWAA17}
Andrew~G. Howard, Menglong Zhu, Bo Chen, Dmitry Kalenichenko, Weijun Wang,
  Tobias Weyand, Marco Andreetto, and Hartwig Adam.
\newblock Mobilenets: Efficient convolutional neural networks for mobile vision
  applications.
\newblock {\em CoRR}, abs/1704.04861, 2017.

\bibitem{Huang_2017_CVPR}
Gao Huang, Zhuang Liu, Laurens van~der Maaten, and Kilian~Q. Weinberger.
\newblock Densely connected convolutional networks.
\newblock In {\em Proceedings of the IEEE Conference on Computer Vision and
  Pattern Recognition (CVPR)}, July 2017.

\bibitem{DBLP:conf/iclr/JiT19}
Ziwei Ji and Matus Telgarsky.
\newblock Gradient descent aligns the layers of deep linear networks.
\newblock In {\em 7th International Conference on Learning Representations,
  {ICLR} 2019, New Orleans, LA, USA, May 6-9, 2019}. OpenReview.net, 2019.

\bibitem{Krizhevsky09learningmultiple}
Alex Krizhevsky.
\newblock Learning multiple layers of features from tiny images.
\newblock Technical report, 2009.

\bibitem{liang2019darts+}
Hanwen Liang, Shifeng Zhang, Jiacheng Sun, Xingqiu He, Weiran Huang, Kechen
  Zhuang, and Zhenguo Li.
\newblock Darts+: Improved differentiable architecture search with early
  stopping.
\newblock {\em arXiv preprint arXiv:1909.06035}, 2019.

\bibitem{liu2019auto}
Chenxi Liu, Liang-Chieh Chen, Florian Schroff, Hartwig Adam, Wei Hua, Alan~L
  Yuille, and Li Fei-Fei.
\newblock Auto-deeplab: Hierarchical neural architecture search for semantic
  image segmentation.
\newblock In {\em Proceedings of the IEEE conference on computer vision and
  pattern recognition}, pages 82--92, 2019.

\bibitem{Liu_2018_ECCV}
Chenxi Liu, Barret Zoph, Maxim Neumann, Jonathon Shlens, Wei Hua, Li-Jia Li, Li
  Fei-Fei, Alan Yuille, Jonathan Huang, and Kevin Murphy.
\newblock Progressive neural architecture search.
\newblock In {\em Proceedings of the European Conference on Computer Vision
  (ECCV)}, September 2018.

\bibitem{DBLP:conf/iclr/LiuSVFK18}
Hanxiao Liu, Karen Simonyan, Oriol Vinyals, Chrisantha Fernando, and Koray
  Kavukcuoglu.
\newblock Hierarchical representations for efficient architecture search.
\newblock In {\em 6th International Conference on Learning Representations,
  {ICLR} 2018, Vancouver, BC, Canada, April 30 - May 3, 2018, Conference Track
  Proceedings}. OpenReview.net, 2018.

\bibitem{liu2018darts}
Hanxiao Liu, Karen Simonyan, and Yiming Yang.
\newblock {DARTS}: Differentiable architecture search.
\newblock In {\em International Conference on Learning Representations}, 2019.

\bibitem{Ma_2018_ECCV}
Ningning Ma, Xiangyu Zhang, Hai-Tao Zheng, and Jian Sun.
\newblock Shufflenet v2: Practical guidelines for efficient cnn architecture
  design.
\newblock In {\em Proceedings of the European Conference on Computer Vision
  (ECCV)}, September 2018.

\bibitem{neyshabur2017pac}
Behnam Neyshabur, Srinadh Bhojanapalli, and Nathan Srebro.
\newblock A pac-bayesian approach to spectrally-normalized margin bounds for
  neural networks.
\newblock {\em arXiv preprint arXiv:1707.09564}, 2017.

\bibitem{NIPS2015_5797}
Behnam Neyshabur, Russ~R Salakhutdinov, and Nati Srebro.
\newblock Path-sgd: Path-normalized optimization in deep neural networks.
\newblock In C. Cortes, N.~D. Lawrence, D.~D. Lee, M. Sugiyama, and R. Garnett,
  editors, {\em Advances in Neural Information Processing Systems 28}, pages
  2422--2430. Curran Associates, Inc., 2015.

\bibitem{DBLP:conf/icml/PhamGZLD18}
Hieu Pham, Melody~Y. Guan, Barret Zoph, Quoc~V. Le, and Jeff Dean.
\newblock Efficient neural architecture search via parameter sharing.
\newblock In Jennifer~G. Dy and Andreas Krause, editors, {\em Proceedings of
  the 35th International Conference on Machine Learning, {ICML} 2018,
  Stockholmsm{\"{a}}ssan, Stockholm, Sweden, July 10-15, 2018}, volume~80 of
  {\em Proceedings of Machine Learning Research}, pages 4092--4101. {PMLR},
  2018.

\bibitem{DBLP:conf/aaai/RealAHL19}
Esteban Real, Alok Aggarwal, Yanping Huang, and Quoc~V. Le.
\newblock Regularized evolution for image classifier architecture search.
\newblock In {\em The Thirty-Third {AAAI} Conference on Artificial
  Intelligence, {AAAI} 2019, The Thirty-First Innovative Applications of
  Artificial Intelligence Conference, {IAAI} 2019, The Ninth {AAAI} Symposium
  on Educational Advances in Artificial Intelligence, {EAAI} 2019, Honolulu,
  Hawaii, USA, January 27 - February 1, 2019}, pages 4780--4789. {AAAI} Press,
  2019.

\bibitem{ILSVRC15}
Olga Russakovsky, Jia Deng, Hao Su, Jonathan Krause, Sanjeev Satheesh, Sean Ma,
  Zhiheng Huang, Andrej Karpathy, Aditya Khosla, Michael Bernstein,
  Alexander~C. Berg, and Li Fei-Fei.
\newblock {ImageNet Large Scale Visual Recognition Challenge}.
\newblock {\em International Journal of Computer Vision (IJCV)},
  115(3):211--252, 2015.

\bibitem{DBLP:conf/iclr/SedghiGL19}
Hanie Sedghi, Vineet Gupta, and Philip~M. Long.
\newblock The singular values of convolutional layers.
\newblock In {\em 7th International Conference on Learning Representations,
  {ICLR} 2019, New Orleans, LA, USA, May 6-9, 2019}. OpenReview.net, 2019.

\bibitem{DBLP:conf/iclr/Shu0C20}
Yao Shu, Wei Wang, and Shaofeng Cai.
\newblock Understanding architectures learnt by cell-based neural architecture
  search.
\newblock In {\em 8th International Conference on Learning Representations,
  {ICLR} 2020, Addis Ababa, Ethiopia, April 26-30, 2020}. OpenReview.net, 2020.

\bibitem{DBLP:conf/iclr/SuzukiAN20}
Taiji Suzuki, Hiroshi Abe, and Tomoaki Nishimura.
\newblock Compression based bound for non-compressed network: unified
  generalization error analysis of large compressible deep neural network.
\newblock In {\em 8th International Conference on Learning Representations,
  {ICLR} 2020, Addis Ababa, Ethiopia, April 26-30, 2020}. OpenReview.net, 2020.

\bibitem{Szegedy_2015_CVPR}
Christian Szegedy, Wei Liu, Yangqing Jia, Pierre Sermanet, Scott Reed, Dragomir
  Anguelov, Dumitru Erhan, Vincent Vanhoucke, and Andrew Rabinovich.
\newblock Going deeper with convolutions.
\newblock In {\em Proceedings of the IEEE Conference on Computer Vision and
  Pattern Recognition (CVPR)}, June 2015.

\bibitem{Tan_2019_CVPR}
Mingxing Tan, Bo Chen, Ruoming Pang, Vijay Vasudevan, Mark Sandler, Andrew
  Howard, and Quoc~V. Le.
\newblock Mnasnet: Platform-aware neural architecture search for mobile.
\newblock In {\em Proceedings of the IEEE/CVF Conference on Computer Vision and
  Pattern Recognition (CVPR)}, June 2019.

\bibitem{DBLP:conf/icml/TangGD17}
Junqi Tang, Mohammad Golbabaee, and Mike~E. Davies.
\newblock Gradient projection iterative sketch for large-scale constrained
  least-squares.
\newblock In Doina Precup and Yee~Whye Teh, editors, {\em Proceedings of the
  34th International Conference on Machine Learning, {ICML} 2017, Sydney, NSW,
  Australia, 6-11 August 2017}, volume~70 of {\em Proceedings of Machine
  Learning Research}, pages 3377--3386. {PMLR}, 2017.

\bibitem{Vapnik1998}
Vladimir~N. Vapnik.
\newblock {\em Statistical Learning Theory}.
\newblock Wiley-Interscience, 1998.

\bibitem{DBLP:conf/iclr/XieZLL19}
Sirui Xie, Hehui Zheng, Chunxiao Liu, and Liang Lin.
\newblock {SNAS:} stochastic neural architecture search.
\newblock In {\em 7th International Conference on Learning Representations,
  {ICLR} 2019, New Orleans, LA, USA, May 6-9, 2019}. OpenReview.net, 2019.

\bibitem{xu2019pc}
Yuhui Xu, Lingxi Xie, Xiaopeng Zhang, Xin Chen, Guo-Jun Qi, Qi Tian, and
  Hongkai Xiong.
\newblock Pc-darts: Partial channel connections for memory-efficient
  differentiable architecture search.
\newblock {\em arXiv preprint arXiv:1907.05737}, 2019.

\bibitem{DBLP:conf/cvpr/ZhangZLS18}
Xiangyu Zhang, Xinyu Zhou, Mengxiao Lin, and Jian Sun.
\newblock Shufflenet: An extremely efficient convolutional neural network for
  mobile devices.
\newblock In {\em 2018 {IEEE} Conference on Computer Vision and Pattern
  Recognition, {CVPR} 2018, Salt Lake City, UT, USA, June 18-22, 2018}, pages
  6848--6856. {IEEE} Computer Society, 2018.

\bibitem{zhou2020theory}
Pan Zhou, Caiming Xiong, Richard Socher, and Steven~CH Hoi.
\newblock Theory-inspired path-regularized differential network architecture
  search.
\newblock {\em arXiv preprint arXiv:2006.16537}, 2020.

\bibitem{DBLP:conf/iclr/ZophL17}
Barret Zoph and Quoc~V. Le.
\newblock Neural architecture search with reinforcement learning.
\newblock In {\em 5th International Conference on Learning Representations,
  {ICLR} 2017, Toulon, France, April 24-26, 2017, Conference Track
  Proceedings}. OpenReview.net, 2017.

\bibitem{zoph2018learning}
Barret Zoph, Vijay Vasudevan, Jonathon Shlens, and Quoc~V Le.
\newblock Learning transferable architectures for scalable image recognition.
\newblock In {\em Proceedings of the IEEE conference on computer vision and
  pattern recognition}, pages 8697--8710, 2018.

\end{thebibliography}
}

\newpage
\renewcommand{\thesection}{\Alph{section}}
\setcounter{section}{0}
\section{Estimation of spectral norm of a convolution}
\label{appendix:power_iteration}
As described in Subsec. \ref{method:setting_spectral_norm}, we constrain the spectral norms of all the convolutions $c_p^v$ contained in the over-parameterized network.
We use a power-iteration algorithm \cite{devries2017improved} described in Algorithm \ref{alg:spectral_norm_estimation} to estimate the spectral norm of each convolution.
\begin{algorithm}[t]
  \caption{Power-iteration algorithm for calculating the spectral norm $\sigma_1$ of convolution $c$}
  \label{alg:spectral_norm_estimation}
  \begin{algorithmic}
    \STATE Randomly initialize input $a_0$
    \FOR{$i=1$ to $Z$}
    \STATE $b_i \leftarrow c(a_{i-1})$
    \STATE Normalize $b_i$
    \STATE $a_i \leftarrow c^\mathsf{T}(b_i)$
    \STATE Normalize $a_i$
    \ENDFOR
    \STATE $\sigma_1(c) \leftarrow \|c(a_i)\|_2$
  \end{algorithmic}
\end{algorithm}
Note that the index $v$ and $p$ of $c_p^v$ are omitted for simplicity.
The spectral norm of convolution $c$ is denoted by $\sigma_1(c)$. 
$c(a)$ is a convolutional calculation that takes input $a$.
$c^\mathsf{T}$ is transposed convolution of $c$.
The Algorithm \ref{alg:spectral_norm_estimation} yields an under-estimate, {\it i.e.}, $\tilde{\sigma}_1(c) \leq \|c\|_2$.
In the experiment, we confirmed all the spectral norms of the convolutions in the over-parameterized network are estimated correctly.
We set the number of iteration $Z$ to $5$.

\section{Comparison with the maximum stable rank architecture}
\label{appendix:compare_max_stable_rank}
We hypothesize the network with lower rank convolutions has higher generalization ability (see Subsec. \ref{method:MSR-DARTS}).
In our pipeline, we select the operators that have lowest stable rank in the mixed edges after training of the over-parameterized network.
We examined the low rank convolution brings better generalization performance by comparing MSR-DARTS with the architecture constructed by selecting the operators that have the maximum stable rank in the mixed edges. 

\begin{figure}[tp]
    \includegraphics[bb=0 0 2024 657, width=1.0\columnwidth]{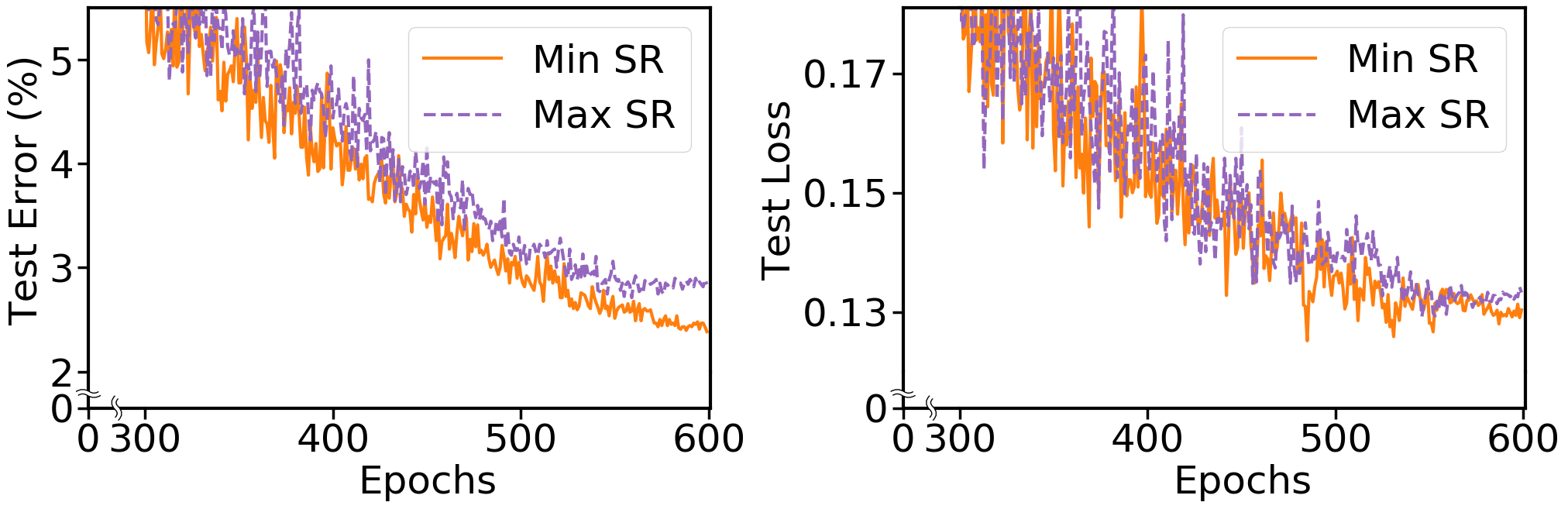}
    \caption{Test-error transition (left) and test-loss transition (right) of the networks generated with MSR-DARTS and the architecture with the highest stable rank operators (after 300 epochs are plotted). Orange solid lines represent MSR-DARTS and purple dotted lines represent the architecture with the highest stable rank operators.}
    \label{fig:compare_stble_rank_max_min} 
\end{figure}
Figure \ref{fig:compare_stble_rank_max_min} shows the results of the training of the both architectures.
We observed the proposed method, {\it i.e.}, the architecture with the lowest stable rank operators yields better performance than the architecture with the highest stable rank operators.
\end{document}